\documentclass[letterpaper]{article} 
\usepackage{aaai2026}  
\usepackage{times}  
\usepackage{helvet}  
\usepackage{courier}  
\usepackage[hyphens]{url}  
\usepackage{graphicx} 
\usepackage{natbib}  
\usepackage{caption} 
\usepackage{algorithm}

\usepackage{newfloat}
\usepackage{listings}
\DeclareCaptionStyle{ruled}{labelfont=normalfont,labelsep=colon,strut=off} 
\floatstyle{ruled}
\newfloat{listing}{tb}{lst}{}
\floatname{listing}{Listing}
\usepackage{graphicx}
\usepackage{array}
\usepackage{amsmath}
\usepackage{bm}
\usepackage{algorithm}
\usepackage{algpseudocode}
\usepackage{amsmath}
\usepackage{multirow}
\usepackage{indentfirst}
\usepackage{subfigure}
\usepackage{epsfig}
\usepackage{epstopdf}
\usepackage{graphicx}
\usepackage{url}
\usepackage{xspace}
\usepackage{booktabs}
\usepackage{subeqnarray}
\usepackage{subcaption}
\usepackage{color}
\usepackage{amssymb}

\newcommand{\paratitle}[1]{\vspace{1.5ex}\noindent\textit{\textbf{#1}}\quad}

\newcommand{\ignore}[1]{}

\title{BEE-RAG: Balanced Entropy Engineering for Retrieval-Augmented Generation}
\author{
    Yuhao Wang\textsuperscript{\rm 1,3}\thanks{~~Equal contributions.}\thanks{~~The work was done during the internship at Baidu.} \quad
    Ruiyang Ren\textsuperscript{\rm 1}\footnotemark[1] \quad
    Yucheng Wang\textsuperscript{\rm 2} \quad
    Jing Liu\textsuperscript{\rm 2}\thanks{~~Corresponding authors.} \\
    Wayne Xin Zhao\textsuperscript{\rm 1,3}\footnotemark[3]  \quad
    Hua Wu\textsuperscript{\rm 2}  \quad
    Haifeng Wang\textsuperscript{\rm 2}
}
\affiliations {
    \textsuperscript{\rm 1}Gaoling School of Artificial Intelligence, Renmin University of China\\
    \textsuperscript{\rm 2}Baidu Inc.\\
        \textsuperscript{\rm 3}Beijing Key Laboratory of Research on Large Models and Intelligent Governance \\
    \{yh.wang500, reyon\_ren\}@outlook.com, batmanfly@gmail.com
}

\usepackage{bibentry}

\begin{document}

\maketitle

\begin{abstract}
With the rapid advancement of large language models~(LLMs), retrieval-augmented generation~(RAG) has emerged as a critical approach to supplement the inherent knowledge limitations of LLMs.
However, due to the typically large volume of retrieved information, RAG tends to operate with extremely long context lengths.
From the perspective of entropy engineering, we identify unconstrained entropy growth and attention dilution due to long retrieval context as significant factors affecting RAG performance.
In this paper, we propose the balanced entropy-engineered RAG~(BEE-RAG) framework, which improves the adaptability of RAG systems to varying context lengths through the principle of entropy invariance. 
By leveraging balanced context entropy to reformulate attention dynamics, BEE-RAG separates attention sensitivity from context length, ensuring a stable entropy level. 
Building upon this, we introduce a zero-shot inference strategy for multi-importance estimation and a parameter-efficient adaptive fine-tuning mechanism to obtain the optimal balancing factor for different settings. 
Extensive experiments across multiple RAG tasks demonstrate the effectiveness of BEE-RAG.

\end{abstract}

\section{Introduction}

Retrieval-augmented generation~(RAG) has emerged as a transformative paradigm for augmenting large language models~(LLMs) with dynamically integrated external knowledge~\cite{wang2025unveiling,lewis2020retrieval}. While conventional RAG systems demonstrate remarkable capabilities in knowledge-intensive tasks, their performance exhibits critical fragility when processing extended contextual inputs~\cite{ren2023investigating}. This vulnerability arises from the inherent requirement of RAG frameworks to incorporate external retrieved documents, which typically results in substantially longer input sequences compared to standard generation tasks~\cite{zhang2024soaring, li2025analyzing}.
This limitation manifests acutely in scenarios demanding long-context comprehension, such as scientific literature analysis, multi-hop reasoning, and cross-document synthesis, where the growing complexity of modern knowledge systems necessitates robust processing of extended text spans~\cite{liu2023lost}.

\begin{figure}[t]
    \centering
    \includegraphics[width=0.46\columnwidth]{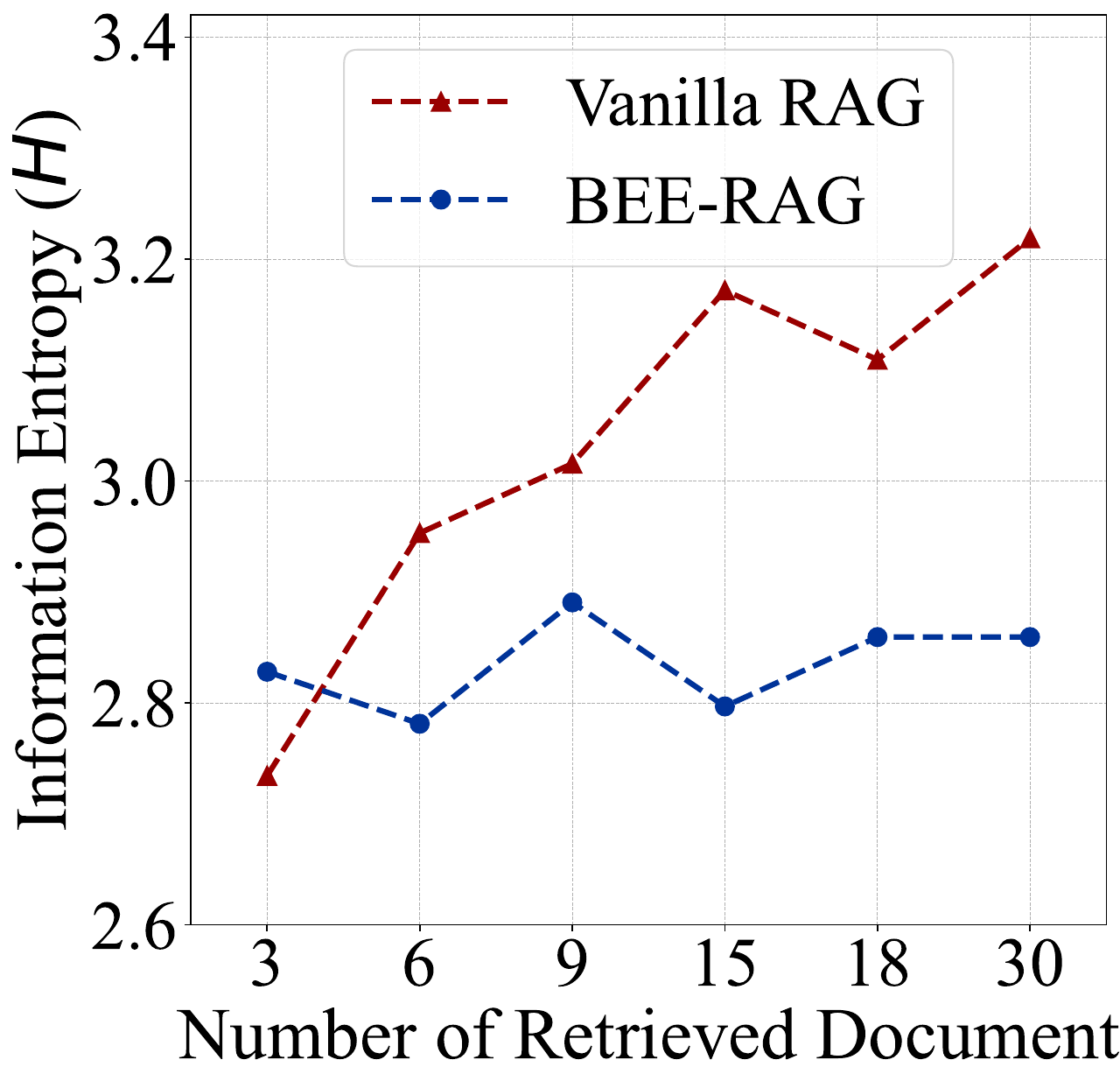} 
    \vspace{2pt} 
    \includegraphics[width=0.51\columnwidth]{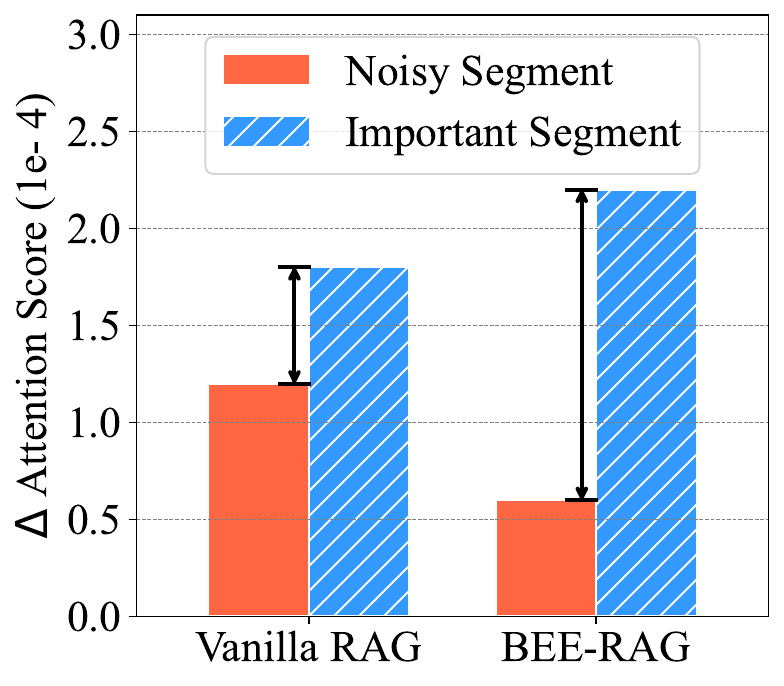} 
    \vspace{-5pt} 
    \caption{In vanilla RAG, increasing input context length raises the information entropy of attention scores (left), potentially harming performance, while the LLM is less focused on important segments (right). This paper proposes a balanced entropy engineering strategy, which maintains entropy stability for longer contexts and guide the LLM to focus on critical segments.}
    \vspace{-2ex}
    \label{fig:motivation}
\end{figure}

Typically, existing efforts to mitigate this challenge focus predominantly on heuristic document filtering, extrapolative training, or architectural modifications, introducing worth noting trade-offs. 
Threshold-based retrieval document truncation or selection reduces context length at the cost of discarding potentially relevant information~\cite{jeong2024adaptive, wang-etal-2024-rear}.
While extended context training may enhance RAG performance~\cite{linra, tang2025enhancing}, it substantially increases computational demands and risks compromising model generalization~\cite{zhan2024over}.
The introduction of auxiliary modules, such as specialized encoders~\cite{DBLP:conf/acl/YenG024}, further escalates the complexity of the system~\cite{ren2025llm}.

As shown in Figure~\ref{fig:motivation}, we identify a fundamental oversight in existing explorations: the failure to address the core tension between context entropy growth and attention allocation dynamics. As retrieved document length increases, unconstrained context entropy expansion progressively dilutes attention distributions, undermining LLMs' capacity to prioritize semantically critical content. Such a phenomenon substantially degrades RAG's ability to extract and utilize salient information.

In this work, we fundamentally rethink RAG's context length adaptability through the perspective of entropy engineering, and propose Balanced Entropy-Engineered RAG~(\textit{BEE-RAG}), a principled framework that establishes entropy-invariant dynamics to decouple attention sharpness from context length.
First, we introduce balanced context entropy, a novel attention reformulation incorporating document-specific balancing entropy factor $\beta$ to enforce entropy invariance across variable context lengths. Theoretical analysis demonstrates this mechanism maintains entropy levels within a stable regime for various context scales, contrasting with conventional entropy scaling.
Building on this foundation, we develop intrinsic multi-importance inference, a zero-shot strategy that derives balancing entropy factors through context aggregation while preserving inter-document independence via parallel context scoring and generation.
To ensure broad applicability, we further propose adaptive balancing factor learning, which is a parameter-efficient tuning method that learns task-specific balancing factor through lightweight linear projections for domain adaptation with minimal parameter updates (0.014\% of total parameters).

Empirical analyses across multiple real-world benchmarks reveal that BEE-RAG fundamentally alters the entropy scaling laws governing conventional RAG systems, which provides a robust and scalable solution for RAG systems dealing with long-context scenarios.
Our main contributions are summarized as follows:
\begin{itemize}
    \item We introduce the concept of balanced context entropy, a novel attention reformulation that ensures entropy invariance across varying context lengths, and allocates attention to important segments. It addresses the critical challenge of context expansion in RAG.
    \item We propose a zero-shot strategy called \textit{intrinsic multi-importance inference} and parametric-efficient fine-tuning strategy adaptive balancing entropy factor learning to obtain the balancing entropy factor in different scenarios while maintaining efficiency.
    \item Extensive empirical validation across multiple benchmarks demonstrates the effectiveness and efficiency of BEE-RAG.

\end{itemize}
\section{Preliminaries}
\label{sec:preliminary}

\paratitle{Task Formulation.}\quad
The task of retrieval-augmented generation~(RAG) involves the integration of retrieval-based techniques with generative models to enhance the accuracy and coherence of text generation tasks~\cite{lewis2020retrieval}. 
A typical RAG system consists of a corpus \( C \), a retriever \( R \), and a generative model \( LLM \). Given a user query or input context \( q \), the retriever \( R \) retrieves relevant documents \( D_\text{q} \) from the corpus.
The generative model \( LLM \) uses both the query \( q \) and the retrieved documents to generate a response:
\begin{equation}
    a = \arg \max_{y} P(y \mid LLM(q, D_q)).
\end{equation}
We focus on how to enhance the multi-document utilization ability of the language model given \( D_q \) and \( q \), with an emphasis on improving the LLM's ability to leverage contextual knowledge, thereby improving its performance on factual generation.

\paratitle{Entropy Engineering.}\quad
In this work, we reconsider RAG from an information-theoretic perspective and introduce the concept of \emph{Entropy Engineering}.
In the standard Transformer~\cite{waswani2017attention}, the attention is computed as:
\begin{equation}
    \text{Attention}(\mathbf{Q}, \mathbf{K}, \mathbf{V}) = \text{Softmax}\left(\frac{\mathbf{QK}^T}{\sqrt{d}}\right) \mathbf{V},
\end{equation}
where $\mathbf{Q}$, $\mathbf{K}$, $\mathbf{V} \in \mathbb{R}^{n \times d}$ represent the query, key, and value matrices, respectively, and the scaling factor \(\sqrt{d}\) stabilizes the dot-product magnitudes. This can be simplified by introducing a scaling parameter \(\lambda = \frac{1}{\sqrt{d}}\):
\begin{equation}
a_{i,j} = \frac{e^{\lambda\, \bm{q_i} \cdot \bm{k_j}}}{\sum_{j=1}^{n} e^{\lambda\, \bm{q_i} \cdot \bm{k_j}}}.
\label{eq:attention}
\end{equation}
We examine this formula from two perspectives:
{From the perspective of token $t_j$ being attended to}: a larger $a_{i,j}$ indicates that the LLM pays more attention to the information at position $i$. Therefore, the LLM should be guided to focus on relatively important information.
{From the perspective of token $t_i$ receiving information}, the values of $\{a_{i,j}\ | i \leq j\}$ across all positions collectively determine how much information $t_i$ receives.

Based on this, inspired by existing analyses~\cite{zhang2024attention}, we measure the amount of information a position receives using discrete entropy, as shown in the following equation:
\begin{equation}
H_i = -\sum_{j=1}^{n} a_{i,j} \log a_{i,j},
\label{eq:entropy}
\end{equation}
which quantifies how much information \( t_i \) receives from the attention perspective. This insight suggests that LLMs struggle with longer sequences when not trained on them, likely due to the discrepancy in information received by tokens in longer contexts. 
Based on the previous analysis, the optimization of attention entropy should focus on two aspects:

\textbullet~The information entropy at positions that are relatively important and likely contain key information should increase.

\textbullet~The overall total information entropy of the context should ideally remain stable with respect to \( n \), to prevent the model from being biased or misaligned when handling longer contexts.

\section{Methodology}
In this section, we first
present a brief overview of our method, then we comprehensively elucidate the methodological details from the perspectives of entropy control and specific zero-shot and parametric-efficient fine-tuning strategies.

\subsection{Overview}
The performance of traditional RAG systems often degrades as the input context length increases, primarily due to the escalating context entropy that dilutes attention distributions across retrieved documents. 
To mitigate this issue, we propose \textbf{B}alanced \textbf{E}ntropy \textbf{E}ngineering for \textbf{RAG}~(\textit{BEE-RAG}), stabilizing context entropy while adaptively optimizing attention allocation to enhance RAG performance. Our method is driven by two key insights: (1) controlling context entropy growth is critical for maintaining robustness in RAG systems with extended contexts, and (2) document importance should dynamically guide attention allocation without costly computation or parametric updates.

The BEE-RAG framework introduces three principal innovations. First, we propose \textit{balanced context entropy}, which incorporates an additive \textit{balancing entropy factor} $\beta$ into attention computation to maintain entropy invariance across varying context lengths. 
Theoretical analysis confirms that applying specific distributional constraints can effectively eliminate the impact of context length on information entropy.
Second, we develop a \textit{intrinsic multi-importance inference} strategy that computes the balancing entropy factor in a zero-shot manner 
, eliminating the need for auxiliary models or training data. 
Third, for scenarios requiring domain adaptation, we devise a parameter-efficient \textit{adaptive balancing entropy factor learning} approach that 
achieves enhanced performance with minimal computational overhead.

\subsection{Entropy Control}
In this section, we first introduce the proposed balancing entropy factor to maintain the stability of context entropy as the input length increases, and then provide a theoretical analysis.

\subsubsection{Balanced Context Entropy}
\label{sec:beta}
As outlined in Section~\ref{sec:preliminary}, we aim to modify the attention mechanism to achieve two critical objectives: (1) maintaining the overall information entropy without a significant increase as a function of sequence length $n$, and (2) ensuring that significant positions receive higher attention scores. This entropy control is designed to enable LLM to maintain superior performance when processing extended contexts by dynamically allocating for attention weights.

Concretely, we propose a new conception {balanced context entropy} from the perspective of entropy engineering to satisfy the above goals by introducing an additive {balancing entropy factor} $\beta \in \mathbb{R}^n$ into the attention computation:
\begin{equation}
    a_{i,j} = \frac{\exp\left(\frac{\mathbf{q}_i \cdot \mathbf{k}_j}{\sqrt{d}} + \beta_{i}\right)}{\sum_{l=1}^n \exp\left(\frac{\mathbf{q}_i \cdot \mathbf{k}_l}{\sqrt{d}} + \beta_{i}\right)}.
\label{eq:bce}
\end{equation}
To focus on important documents, we set all tokens of the same chunk to share the same value. In this case, the balancing entropy factor \(\beta_i\) represents the importance of an individual retrieved document according to the query, with more important documents receiving higher scores. This effectively concentrates the LLM's attention on important documents.

Building upon the proposed conception of BCE, we present two distinct approaches for deriving the balancing entropy factor $\beta$ and systematically elaborate on their methodologies in Section~\ref{sec:framework}.
Before that, we conduct a rigorous theoretical analysis of BCE's operational mechanisms to establish the theoretical rationale for its mathematical formulation.

\subsubsection{Theoretical Analysis}
\label{sec:proof}
Here, we present a theoretical derivation to substantiate the rationale behind the proposed balanced context entropy and discuss how to set constraints on the balancing entropy factor for different context lengths.

We denote \(\xi_{i,j} = \frac{q_i \cdot k_j}{\sqrt{d}}\), and combine Equation~(\ref{eq:attention}), (\ref{eq:entropy}), and (\ref{eq:bce}) to obtain:
\begin{equation}
    H_i = \log \sum_{j=1}^{n} e^{\xi_{i,j} + \beta_i} - \frac{\sum_{j=1}^{n} e^{\xi_{i,j} + \beta_i} ( \xi_{i,j} + \beta_i)}{ \sum_{j=1}^{n} e^{\xi_{i,j} + \beta_i}}
\end{equation}
We assume that each row of the query matrix $\mathbf Q\in\mathbb R^{n\times d}$ and key matrix $\mathbf K\in\mathbb R^{n\times d}$ is an independent, \emph{isotropic sub-Gaussian} vector with zero mean and identity covariance, i.e.,
$\mathbb{E}[\mathbf{q}_i] = \mathbb{E}[\mathbf{k}_i] = \mathbf{0}$.
By the high-dimensional central limit theorem~\cite{bolthausen2013bernoulli}, the scaled dot product 
$\xi_{i,j} = \frac{\mathbf{q}_i^\top \mathbf{k}_j}{\sqrt{d}}$ converges in distribution to a standard Gaussian, i.e., $\xi \xrightarrow{d} \mathcal{N}(0, 1)$.
We further assume that the document-level importance bias follows a Gaussian distribution $\beta \sim \mathcal{N}(\mu, \sigma^2)$ and is independent of $\xi$. This independence reflects the fact that the same token may appear in documents of varying importance, and thus the semantic similarity (captured by $\xi$) and document relevance (captured by $\beta$) should be modeled as separate random variables.
Thus, we have:

\begin{align}
    H_i \approx 
    & \log n + \log \mathbb{E} \left(e^{\xi}\right) + \log \mathbb{E}\left(e^\beta\right) - \nonumber \\
    & \frac{\mathbb{E}\left(\xi e^{\xi}\right) + \mathbb{E}\left(\beta e^{\xi}\right) + \mathbb{E}\left(\xi e^\beta\right) + \mathbb{E}\left(\beta e^\beta\right)}{\mathbb{E}\left(e^{\xi}\right)\mathbb{E}\left( e^\beta\right)},
\end{align}
After simplification, we obtain:
\begin{align}
    H_i \approx 
    & \log n + \frac{1}{2} + \mu + \frac{\sigma^2}{2} - \nonumber \\
    & \frac{e^{\frac{1}{2}} + \mu e^{\frac{1}{2}} + \left(\mu + {\sigma^2}\right) e^{\mu + \frac{\sigma^2}{2}}}{e^{\mu+\frac{\sigma^2}{2}+\frac{1}{2}}}.
\end{align}
We aim for entropy to remain invariant as the context length \( n \) increases, thereby ensuring that the performance of RAG remains unaffected~(seen in Section~\ref{sec:preliminary}). Thus, \(\mu\) and \(\sigma\) should satisfy the following formulation:
\begin{align}
    -\log n = \mu + \frac{\sigma^2}{2} -
    \frac{e^{\frac{1}{2}} + \mu e^{\frac{1}{2}} + \left(\mu + {\sigma^2}\right) e^{\mu + \frac{\sigma^2}{2}}}{e^{\mu+\frac{\sigma^2}{2}+\frac{1}{2}}}.
\label{eq:h-const}
\end{align}

Based on the solution to Equation~(\ref{eq:h-const}), we need to constrain \(\sigma\) to increase as \(n\) grows, ensuring that the context entropy remains stable as the context length increases. Our experimental results confirm that the parameters derived from theory are consistent with those selected empirically.

\subsection{BEE-RAG Framework}
\label{sec:framework}

\begin{figure*}[ht!]
  \centering
  \includegraphics[width=0.89\textwidth]{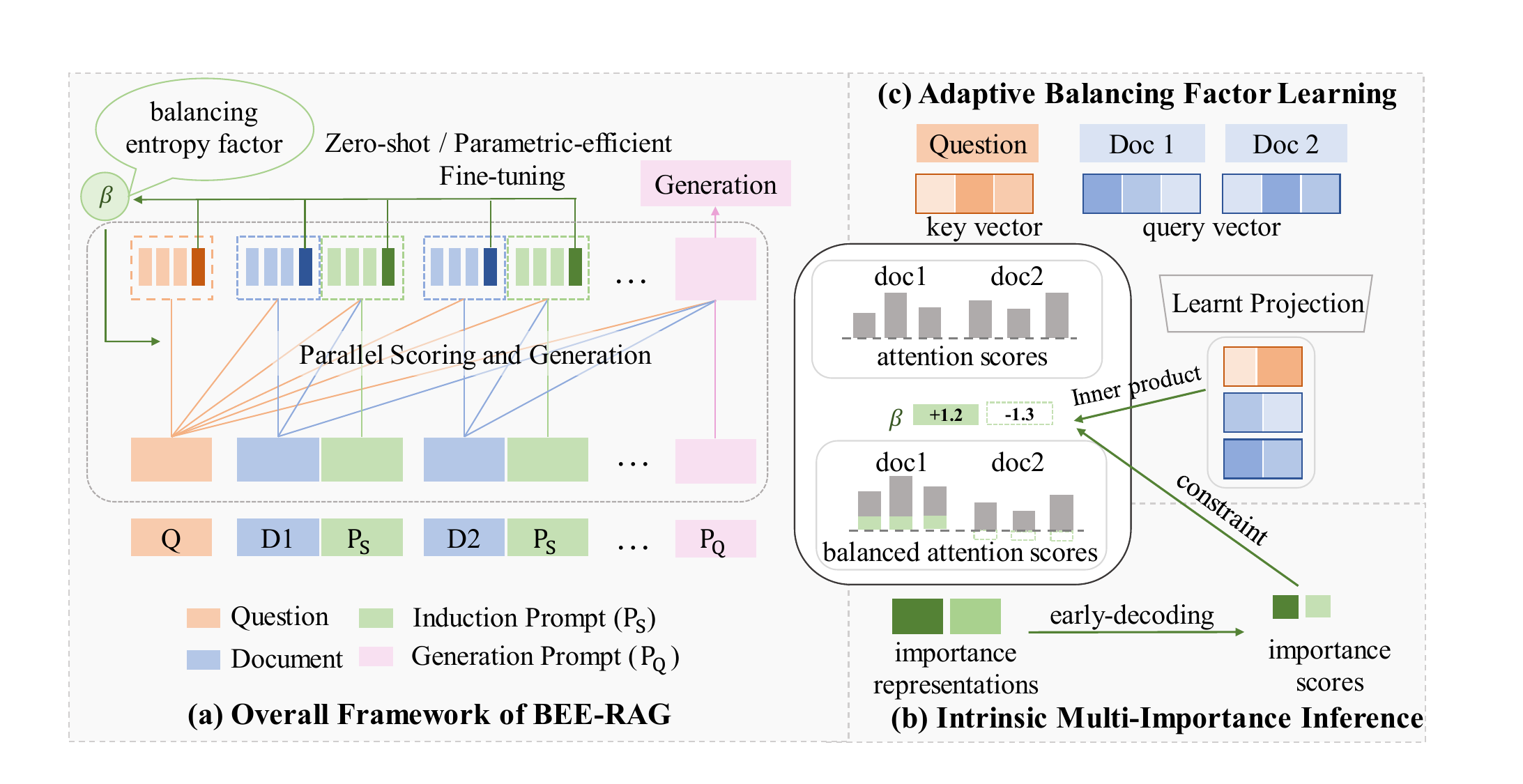}
  \caption{The overview of the proposed BEE-RAG. The left side shows the overall architecture design, while the right side illustrates the zero-shot strategy \textit{intrinsic multi-importance inference} and the parametric-efficient fine-tuning strategy \textit{adaptive balancing factor learning}.}
  \label{fig:framework}
\end{figure*}

In the preceding section, we introduced balanced context entropy as a methodological enhancement to stabilize the performance of RAG systems. We hereby propose efficient methods to determine the balancing entropy factor for both zero-shot learning and parameter-efficient fine-tuning.

\subsubsection{Intrinsic Multi-Importance Inference}
\label{sec:imi}
In this part, we introduce the intrinsic multi-importance inference (IMI) approach for zero-shot reasoning in BEE-RAG. 
The core idea of IMI lies in leveraging the intrinsic parameters of LLM to assess the importance of individual documents within RAG inputs as the balancing entropy factor in BCE computation, thereby eliminating the need for external modules while maintaining computational efficiency and architectural simplicity.

Inspired by the success of prompt engineering for LLMs in retrieval and ranking tasks~\cite{zhuang2024setwise, pradeep2023rankvicuna}, we use a prompt-induced attention calibration strategy to obtain importance scores for each document. Specifically, we append a prompt to each document for obtaining hidden state representations of importance. To extract importance scores from these representations, we apply an early-decoding strategy~\cite{cammarata2020thread:}, using lm\_head(·) to map the generation probability of a critic token as the document's importance score at each layer. For instance, the prompt can be ``\textit{Does the passage support the answer to the question?}'', the critic token is ``\textit{yes}'').

Since the characteristics of LLM, the derived importance scores show individual distributions in scales and dispersion. To enable stable entropy control across contexts and generalize to different context lengths, we constrain the distribution to be stable with consistent mean and variance, based on the formulation in Equation~\ref{eq:h-const}. Concretely, we normalize and rescale the importance scores to a target mean \(\mu\) and variance \(\sigma^2\).
This distributional alignment ensures comparability of importance while preserving the intrinsic ranking relationships between documents.

The cross-attention mechanisms employed in LLMs may inadvertently influence importance score estimation for individual documents, as it enables inter-document interference through shared attention parameters.
Inspired by previous work~\cite{ratner2023parallel}, we develop a parallel scoring mechanism in RAG architecture through attention mask modifications. This enables simultaneous computation of document importance and task outputs in a single forward pass.
In this way, we create independent processing streams that prevent cross-document contamination, which is a critical foundation for stabilizing the context entropy. 

\subsubsection{Adaptive balancing entropy factor Learning}
\label{sec:train}
To extend our entropy engineering framework to domain-specific scenarios, we introduce an adaptive balancing entropy factor learning strategy with improved training efficiency. we first introduce designed vector fine-tuning for sentence-level importance scoring. Then, we implement initialization constraints to ensure stable training convergence.

The core of our approach lies in the specially designed efficient vector fine-tuning method, which bridges token-level and sentence-level representations. Specifically, inspired by previous work~\cite{wu2025reft}, we transforms token-level vectors into sentence-level vectors:
\begin{equation}
\bm{v_\text{sentence}} = \bm{v_b} + \bm{v_e} + \bm{R}^\intercal (\bm{Rv_b} - \bm{Rv_e}),
\end{equation}
where $\bm{v}$ represents either key or query vectors, and $\bm{v_b}$, $\bm{v_e}$ correspond to the start and end token vectors of documents or queries, respectively. The $\bm{R} \in \mathbb{R}^{d \times d_r}$ a trainable projection matrix with $d_r \ll d$. With the sentence-level vectors, we can calculate the importance scores through query-key inner products:
\begin{equation}
    \bm{\beta} = \bm{q_\text{doc}} \cdot \bm{k_\text{query}}.
\end{equation}
During implementation, we observe unconstrained values of $\bm{\beta}$ leading to gradient explosion. To address it, we employ orthogonal initialization~\cite{saxe2013exact} on \(\bm{R}\) and implement additional scaling on $\bm{\beta}$ when its values exceed a certain threshold.

\subsection{Discussion}
\label{subsec:discuss}

\paratitle{Novelty and Effectiveness.}
{
Our key innovation lies in introducing an additive balancing entropy factor into the attention mechanism, achieving two critical objectives: (1) maintaining stable context entropy across varying input lengths through mean and variance constraints, thereby enhancing performance in long-context scenarios; and (2) adaptively guiding the LLM to focus on crucial segments, improving its ability to effectively utilize knowledge. Unlike existing attention reweighting approaches that either uniformly reduce context entropy~\cite{kexuefm-8823} (hindering important information extraction) or ignore attention entropy~\cite{yan2024corrective} (leading to unstable performance across varying context lengths). We further introduce novel adaptations for zero-shot and parameter-efficient fine-tuning scenarios: a prompt-induced importance scoring mechanism with parallel scoring-generation architecture, and an efficient vector fine-tuning strategy. These components collectively enhance BEE-RAG's accuracy in question-answering tasks
}

\paratitle{Efficiency.}
{
We further analyze the efficiency of our EE-RAG framework. The parallel mechanism handles \(\beta\) calculation and answer generation simultaneously. Due to the quadratic time complexity in the transformer mechanism, our design effectively controls computational overhead, achieving performance comparable.}
\section{Experiments}
In this section, we detail the experimental setup
and then report the evaluation results.

\subsection{Experimental Settings}

\subsubsection{Datasets} 
We conduct experiments on both single-hop and multi-hop question-answering (QA) tasks. Specifically, we evaluate our approach on the Natural Questions~(NQ)~\cite{kwiatkowski2019natural}, TriviaQA~\cite{joshi2017triviaqa}, HotpotQA~\cite{Yang2018HotpotQAAD}, and 2WikiMultihopQA~(2WikiQA)~\cite{ho2020constructing} datasets.

\subsubsection{Metrics}
We use Exact Match (\textbf{EM})~\cite{lee2019latent} and \textbf{F1}~\cite{dpr2020} to assess the QA accuracy of retrieval-augmented LLMs, which are commonly adopted in QA evaluation. EM checks if predicted answers exactly match the ground truth, while F1 measures their overlap in precision and recall.

\subsubsection{Implementation Details}
We select LLaMA-3-8B~\cite{dubey2024llama} and Qwen-2.5-7B~\cite{yang2024qwen2} as base LLM. For the zero-shot setting (Section~\ref{sec:imi}), we derive the theoretical values for $\mu$ and $\sigma^2$ based on Equation~\ref{eq:h-const}, and perform a search around these values to determine their optimal configuration. For the adaptive balancing entropy factor learning, we set $d_r = 8$. We conduct the experiments on 8 NVIDIA A100 40GB GPUs, with a learning rate of 5e-4, a cosine learning rate scheduler, and a batch size of 16. To demonstrate the robustness of our training approach, we train the LLMs for no more than 500 steps solely on the NQ dataset and subsequently validate its performance across multiple datasets.

\subsubsection{Baselines}
We introduce two types of baselines for comparison with our approach, including zero-shot methods and lightweight fine-tuning methods.

\textbullet~\emph{Zero-shot methods.} We select several prompt-based strategies that enhance generation performance in RAG scenarios for comparison, including prompt-based methods (Chain-of-Thought~\cite{wei2022chain}, Chain-of-Note~\cite{yu2023chain}, Self-Critic~\cite{asaiself}), architecture-driven methods (PCW~\cite{ratner2023parallel}, Position-Engineering~\cite{he2024position}), and attention reweighting method (Multiply-Attention~\cite{chiang2022overcoming}).

\textbullet~\emph{Parametric-efficient fine-tuning methods:} We choose two commonly used PEFT approaches in RAG, including LoRA~\cite{hulora} and Prefix-Tuning~\cite{li2021prefix} for comparison with our parameter-efficient training strategy.

\subsection{Main Results}

\begin{table*}[ht!]
\centering
\small
\renewcommand\tabcolsep{8.0pt}
\scalebox{1.025}{
\begin{tabular}{lcccccccccc}
    \toprule
     \multirow{3}{*}{\textbf{Settings}} & \multicolumn{2}{c}{\textbf{NQ}} & \multicolumn{2}{c}{\textbf{TriviaQA}} & \multicolumn{2}{c}{\textbf{HotpotQA}} & \multicolumn{2}{c}{\textbf{2WikiQA}} & \multicolumn{2}{c}{\textbf{Average}}\\
     \cmidrule(lr){2-3} \cmidrule(lr){4-5} \cmidrule(lr){6-7} \cmidrule(lr){8-9} \cmidrule(lr){10-11} 
      & \textbf{EM} & \textbf{F1} & \textbf{EM} & \textbf{F1} & \textbf{EM} & \textbf{F1} & \textbf{EM} & \textbf{F1} & \textbf{EM} & \textbf{F1}\\
    \midrule
    \multicolumn{11}{c}{\emph{Qwen-2.5-7B}} \\
     \midrule
      Vanilla RAG & 26.84 & 42.03 & 43.05 & 53.41 & 24.54 & 35.16 & 20.16 & 27.16 & 28.65 & 39.44 \\
      Chain-of-Thought & 26.86 & 40.20 & 39.59 & 51.78 & 26.20 & 35.88 & 22.93 & 26.58 & 28.89 & 38.61 \\
      Chain-of-Note & 26.50 & 41.75 & 41.20 & 52.03 & 26.74 & 35.95 & 19.93 & 26.58 & 28.59 & 39.08 \\
      PCW & 23.87 & 38.21 & 40.88 & 51.50 & 24.72 & 34.63 & 23.11 & 27.19 & 28.15 & 37.88 \\
      Position-Engineering & 25.76 & 38.61 & 41.38 & 52.30 & 27.77 & 36.56 & 22.69 & 26.96 & 29.40 & 38.60 \\
      Self-Critic & 24.27 & 35.36 & 40.74 & 52.04 & 25.11 & 35.35 & 22.00 & 27.51 & 28.03 & 37.57 \\
      Multiply-Attention & 26.18 & 41.43 & 42.92 & 53.62 & 27.13 & 34.94 & 18.87 & 24.02 & 28.78 & 38.50 \\
      \textbf{Zero-BEE~(ours)} & \underline{29.03} & \underline{43.15} & \underline{44.32} & \underline{55.93} & \underline{27.86} & \underline{37.15} & \underline{24.47} & \underline{29.61} & \underline{31.42}$^{\dagger}$ & \underline{41.46}$^{\dagger}$ \\
     \midrule
      LoRA~(2.06M) & 37.62 & 47.93 & 47.98 & 56.32 & 28.44 & 38.92 & 20.76 & 26.30 & 33.70 & 42.37 \\
      Prefix-Tuning~(2.07M) & 36.75 & 46.33 & 48.21 & 56.32 & 31.19 & 40.15 & 23.69 & 29.19 & 34.96 & 43.00 \\
      \textbf{Light-BEE~(ours, 2.06M)} & \textbf{38.71} & \textbf{50.11} & \textbf{49.39} & \textbf{56.73} & \textbf{35.95} & \textbf{44.81} & \textbf{28.32} & \textbf{31.73} & \textbf{38.09$^{\dagger}$} & \textbf{45.84$^{\dagger}$} \\
    \midrule
   \multicolumn{11}{c}{\emph{LLaMA-3-8B}} \\
    \midrule
      Vanilla RAG & 27.13 & 37.77 & 41.17 & 55.46 & 23.07 & 34.22 & 14.20 & 23.74 & 26.39 & 37.80 \\
      Chain-of-Thought & 29.10 & 38.60 & 45.31 & 56.96 & 22.74 & 30.23 & 14.50 & 25.58 & 27.91 & 37.84 \\
      Chain-of-Note & 28.46 & 37.83 & 41.49 & 51.22 & 21.34 & 28.99 & 19.87 & 29.50 & 27.79 & 36.89 \\
      PCW & 25.71 & 35.47 & 38.72 & 48.99 & 17.14 & 25.92 & 17.88 & 26.65 & 24.86 & 34.26 \\
      Position-Engineering & 25.98 & 35.72 & 41.01 & 54.21 & 23.24 & 30.62 & 16.08 & 26.61 & 26.58 & 36.79 \\
      Self-Critic & 20.86 & 27.06 & 42.61 & 51.06 & 23.77 & 31.21 & 15.50 & 25.94 & 25.69 & 33.82 \\
      Multiply-Attention & 27.08 & 36.33 & 43.40 & 55.05 & 21.27 & 28.24 & 11.87 & 23.74 & 25.91 & 35.84 \\
      \textbf{Zero-BEE~(ours)} & \underline{30.50} & \underline{38.96} & \underline{45.67} & \underline{57.04} & \underline{23.84} & \underline{34.29} & \underline{20.67} & \underline{30.56} & \underline{30.17}$^{\dagger}$ & \underline{40.21}$^{\dagger}$ \\
      \midrule
      LoRA~(2.62M) & 35.39 & 48.68 & 53.65 & 58.55 & 33.15 & 42.15 & 18.91 & 28.67 & 35.27 & 44.51 \\
      Prefix-Tuning~(2.63M) & 36.39 & 48.25 & 48.75 & 56.48 & 32.30 & 42.50 & 16.33 & 25.13 & 33.44 & 43.09 \\
      \textbf{Light-BEE~(ours, 2.62M)} & \textbf{38.42} & \textbf{49.23} & \textbf{53.96} & \textbf{58.67} & \textbf{38.06} & \textbf{44.72} & \textbf{24.70} & \textbf{31.89} & \textbf{38.79$^{\dagger}$}
 & \textbf{46.13$^{\dagger}$} \\
    \bottomrule
\end{tabular}}
\caption{Main results of BEE-RAG on four RAG tasks with different LLM backbones for both zero-shot inference and lightweight fine-tuning settings. Zero-BEE denotes the zero-shot inference version of BEE-RAG, while Light-BEE denotes the parameter-efficient (lightweight) fine-tuning version. All results are averaged over 8 runs; † indicates statistical significance with $p < 0.05$.}

\label{tab:overall}
\end{table*}

Table~\ref{tab:overall} presents the results of our approach and the baselines on four RAG tasks of open-domain QA. 

First, our proposed method outperforms all other baselines in terms of average QA performance. Whether in the zero-shot setting or with lightweight fine-tuning, BEE-RAG consistently achieves the best results. This highlights the effectiveness of our key motivation: the reduction of context entropy and dynamic attention allocation on documents, which proves to be applicable to both single-hop and multi-hop datasets.

Second, our method demonstrates its ability to handle more challenging datasets, assisting LLMs in effectively addressing complex RAG tasks. On relatively more difficult multi-hop datasets, BEE-RAG shows significant performance improvements. Notably, on the latest 2WikiMultihopQA dataset, our approach achieves an approximately 5\% improvement over vanilla RAG, further highlighting the effectiveness of our approach.

Third, our parametric-efficient fine-tuning method demonstrates better generalization performance with few parameters, emphasizing the effectiveness of the proposed adaptive balancing entropy factor learning approach. The result shows that whether on trained or zero-shot datasets, our method outperforms the others. Notably, the performance of the other two methods may degrade when transferred to multi-hop tasks, revealing their limitations in handling challenging RAG tasks.

\subsection{Ablation Study}

\begin{table}[t]
\centering
\small
\scalebox{1.05}{
\begin{tabular}{lcc|lc}
\toprule
\textbf{Zero-shot} & \textbf{EM} & &\textbf{PEFT} & \textbf{EM}\\
\midrule
\textbf{Zero-BEE} & \textbf{31.42} && \textbf{Light-BEE} & \textbf{38.09}\\
\midrule
w/o $\text{IMI}$ & 29.89 && w/o $\text{IMI}$ & 36.58 \\
w/o $\mu$ & 27.47 && w/o Residual & 23.65\\
w/o $\sigma^2$ & 26.92 && w/o Init constraint & 3.23  \\
w/o $\beta$ & 20.35 && w/ ~ FFN  & 36.41 \\
w/ ~ reranker & 28.87 && w/ ~ reranker & 35.82  \\
\bottomrule
\end{tabular}}
\caption{Ablation study on our BEE-RAG.}
\label{tab:ablation}
\end{table}

In this section, we conduct an ablation study to validate the effectiveness of the key strategies in BEE-RAG on both zero-shot and fine-tuning scenarios. As shown in Table~\ref{tab:ablation}, the left part of the table corresponds to the variants in the zero-shot setting, while the right part focuses on the variants with lightweight fine-tuning.

First, we observe that setting $\mu$ and $\sigma$ to zero (\textit{w/o $\mu$}, \textit{w/o $\sigma^2$}) results in performance degradation, demonstrating that adjusting the mean ($\mu$) and variance ($\sigma^2$) of the balancing entropy factor helps improve multi-document RAG performance.
This observation also aligns well with our formulation in Equation~(\ref{eq:h-const}).

Second, we validate the effectiveness of the specific strategies introduced in our lightweight training approach. We demonstrate the positive effect of our chunk-level adaptive balancing entropy factor learning formula by removing the residual and FFN module in our approach (\textit{w/o Residual}, \textit{w/ FFN}). Furthermore, we prove that adding constraints in the initialization of the balance factor (\textit{w/o Init constraint}) prevents gradient explosion during training.

Third, we observe that removing parallel multi-importance inference (\textit{w/o IMI}) leads to a performance drop, which can be attributed to the interference from other chunks~(documents) when the LLM generates balance factors required independent document information, causing a performance decline. 
In addition, we explore the impact of adding a reranker module to assess the relevance between the queries and documents as a substitute for the balancing entropy factor in BEE-RAG~(w/ reranker). We observe a performance decline, proving our method avoids extra modules and costs while outperforming in both scenarios.

\subsection{In-depth Analysis}
In this section, we further present an analysis from two aspects: the scaling of LLMs and the quality of the documents.

\subsubsection{Effect of Model Scaling}

\begin{figure}[t!]
  \centering
  \includegraphics[width=0.86\columnwidth]{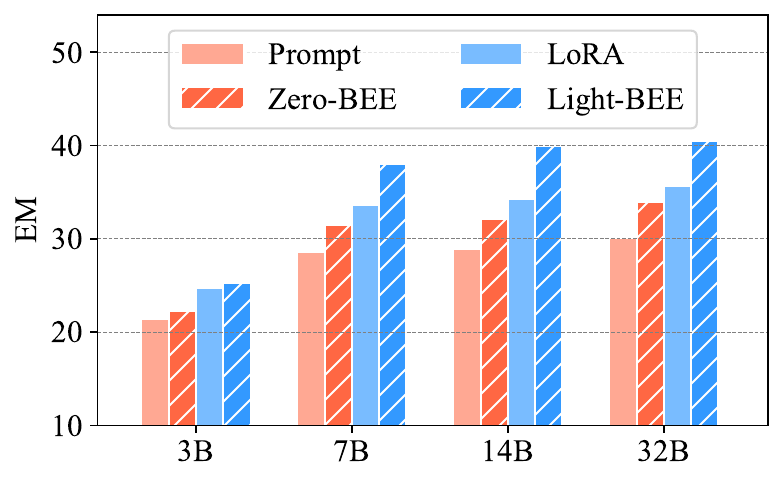}
  \vspace{-0.5ex}
  \caption{Effect of LLM capabilities with various Scales. Mean metrics are reported over four datasets.}
  \label{fig:scale}
\end{figure}

In this part, we first explore the impact of different model scales on experimental performance. We test the Qwen-2.5 family LLMs, ranging from 3B, 7B, 14B to 32B, on their average generation results across four datasets. The results are shown in Figure~\ref{fig:scale}. It is clear that our proposed BEE-RAG consistently improves RAG accuracy across different model sizes. This further demonstrates that our method enables LLMs to better handle generation tasks involving noisy contexts and exhibits strong generalization capabilities.

\begin{figure}[t!]
  \centering
  \includegraphics[width=0.8\columnwidth]{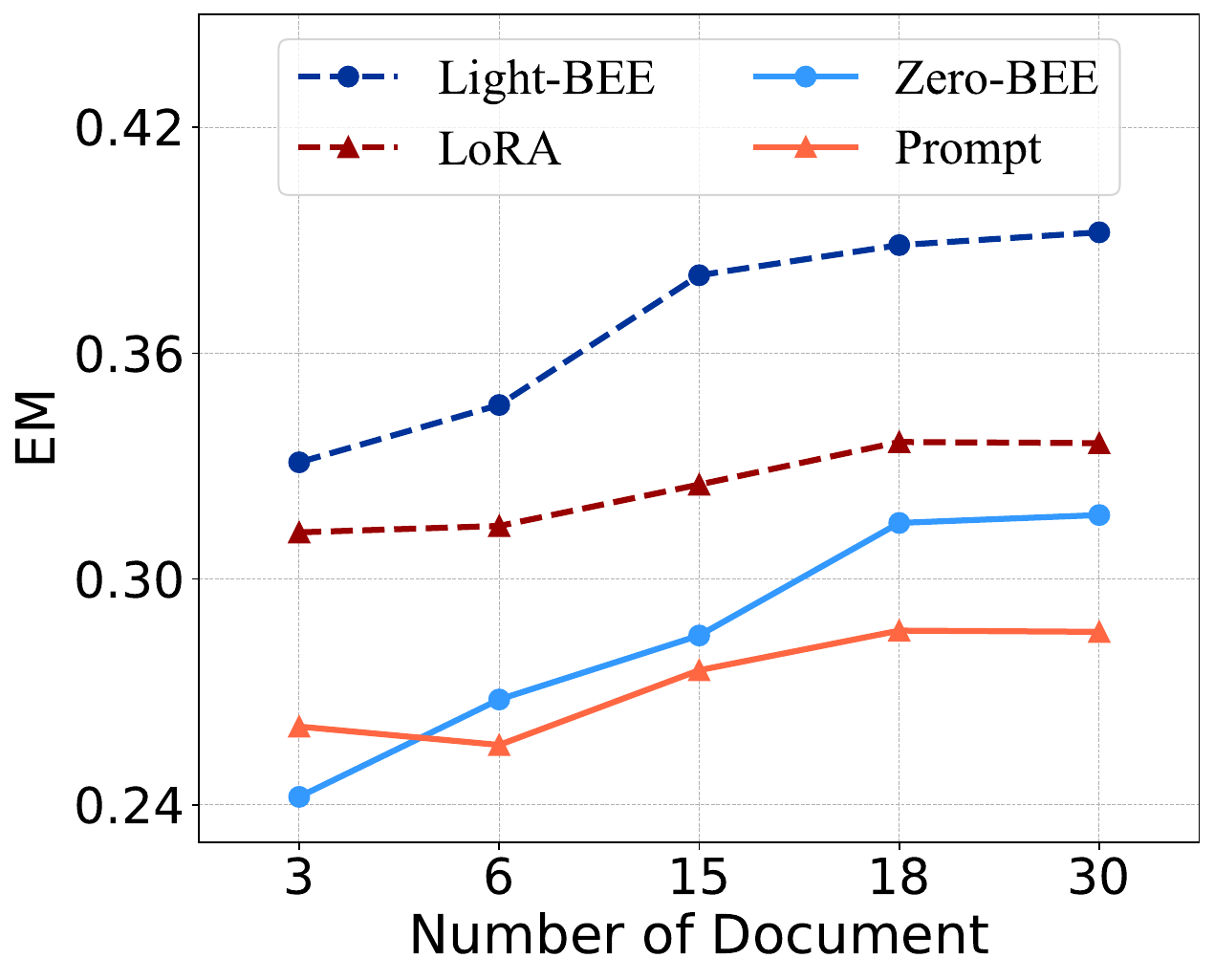}
  \caption{Effect of context length with various document numbers. Mean metrics are reported over four datasets.}
  \label{fig:length}
  \vspace{-2.5ex}
\end{figure}

\subsubsection{Effect of Retrieved Documents}
We consider both context length and retrieval quality to evaluate the effect of retrieved documents, showing the comprehensive priority of the proposed BEE-RAG method.

\paratitle{Context Length.}
We first examine the impact of the number of the input document. Figure~\ref{fig:length} shows the average performance of our BEE-RAG and the baselines varying the document number. It is clear that our method overall outperforms the baselines. Notably, as the context length increases, the performance gap between BEE-RAG and the baselines becomes significantly larger. This highlights that our strategy is more effective in adapting to larger amounts of contextual information, allowing LLMs to efficiently extract important information through global context entropy reduction and dynamic document-level attention allocation.

\begin{table}[t]
\centering
\small
\renewcommand\tabcolsep{4.7pt}
\scalebox{1.05}{
\begin{tabular}{lcccc}
\toprule
\textbf{Retriever} & \textbf{Prompt} & \textbf{Zero-BEE} & \textbf{LoRA} & \textbf{Light-BEE}\\
\midrule
BM25 & 23.82 &  26.98 & 28.22 & 35.03 \\
Contriever & 28.65 & 31.42 & 33.70 & 38.09 \\
E5-Base & 29.36 & 32.59 & 34.92 & 38.55 \\
E5-large & 29.77 & 34.02 & 34.02 & 38.84 \\
BGE & 30.51 & 33.91 & 36.72 & 39.05 \\
\bottomrule
\end{tabular}}
\caption{Effect of retrieval quality with various retrievers. Mean metrics are reported over four datasets.}
\label{tab:retrieve}
\end{table}

\paratitle{Retrieval Quality.}
Moreover, we explore the impact of various retrievers with different retrieval document qualities for the input of BEE-RAG. Table~\ref{tab:retrieve} presents the QA performance of our approach and the baselines using various retrievers across four datasets, including BM25, Contriever, E5-base, E5-large, and BGE. From the table, it is evident that our method achieves the best performance across all retriever settings. Notably, our approach excels when retrieval quality is lower. This highlights that BEE-RAG is better at focusing on the critical parts of the retrieval results, thereby improving the overall accuracy of the generations.

\section{Related Work}

Recent efforts have aimed to improve RAG performance under long contexts~\cite{laban2024summary, wang2025reinforced}. PCW~\cite{ratner2023parallel} restricts attention within fixed-size windows to reduce attention dilution. CEPE~\cite{DBLP:conf/acl/YenG024} uses a lightweight encoder to process inputs chunk by chunk, enabling better use of long contexts via cross-attention. RAFT~\cite{zhang2024raft} and Chain-of-Notes~\cite{yu2023chain} extend the context capacity of models through instruction tuning on long-input data. However, these methods either yield limited performance gains or incur high training costs. Some works explore the effect of attention distribution~\cite{wang2023label}. \citet{kexuefm-8823} adjust attention scaling by context length to match short-context behavior, but this fails to highlight key segments.

Our BEE-RAG introduces an attention reallocation mechanism. It reduces attention entropy gap with short contexts and guides focus to key segments. This improves long-context RAG performance.
\section{Conclusion}

In this study, we identify and address the fundamental limitation of unconstrained context entropy in RAG systems through the lens of information entropy. We proposed BEE-RAG for the adaptability of RAG systems to various context lengths. By introducing balanced context entropy, we demonstrate that enforcing entropy invariance enables superior RAG performance with various context lengths without sacrificing computational efficiency. In order to extend applicability to a broader range of scenarios, we propose multi-importance inference for zero-shot reasoning, and also propose adaptive balancing factor learning for parameter-efficient fine-tuning.
Empirical examination on four real-world knowledge-intensive tasks shows consistent performance gains of BEE-RAG.


\bibliography{aaai2026}



\end{document}